\documentclass{article}

\usepackage{graphicx}
\usepackage{amsmath}
\DeclareMathOperator*{\E}{\mathbb{E}}

     \usepackage[preprint]{neurips_2023}

\usepackage[utf8]{inputenc} % allow utf-8 input
\usepackage[T1]{fontenc}    % use 8-bit T1 fonts
\usepackage{hyperref}       % hyperlinks
\usepackage{url}            % simple URL typesetting
\usepackage{booktabs}       % professional-quality tables
\usepackage{amsfonts}       % blackboard math symbols
\usepackage{microtype}      % microtypography
\usepackage{xcolor}         % colors

\title{A Novel Sparse Regularizer}

\author{%
  Hovig Tigran Bayandorian \\
  \texttt{htbayandorian@gmail.com} 
}

\begin{document}

\maketitle

\begin{abstract}
$L_p$-norm regularization schemes such as $L_0$, $L_1$, and $L_2$-norm regularization and $L_p$-norm-based regularization techniques such as weight decay, LASSO, and elastic net compute a quantity which depends on model weights considered in isolation from one another. This paper introduces a regularizer based on minimizing a novel measure of entropy applied to the model during optimization. In contrast with $L_p$-norm-based regularization, this regularizer is concerned with the spatial arrangement of weights within a weight matrix. This novel regularizer is an additive term for the loss function and is differentiable, simple and fast to compute, scale-invariant, requires a trivial amount of additional memory, and can easily be parallelized. Empirically this method yields approximately a one order-of-magnitude improvement in the number of nonzero model parameters required to achieve a given level of test accuracy when training LeNet300 on MNIST.
\end{abstract}

\section{Introduction}

Deep neural network models \cite{lecun2015deep} provide state-of-the art results on a great variety of learning tasks ranging from wavefunction approximation in many-body quantum physics \cite{pfau2020ab} to the challenging board game of Go \cite{silver2018general}. State-of-the art language models may reach upwards of 1.6 trillion parameters \cite{fedus2021switch} with larger parameter counts associated with better model performance including higher sample efficiency during the learning process \cite{brown2020language}. Despite routinely being initialized and trained in a dense manner, neural networks can be pruned by upwards of 90 percent via a variety of techniques without significantly damaging model accuracy \cite{frankle2018lottery}, but as of yet there is no satisfactory sparsifying regularization technique for model training. When applied to the solution of partial differential equations \cite{brunton2023machine}, sparsification of a deep learning model is a key step for improving the interpretability and generalizability in models of plasma physics  \cite{alves2022data} and of fluid turbulence   \cite{beetham2020formulating}.

An overview of techniques for inducing sparsification in neural networks can be found in \cite{hoefler2021sparsity}, including a review of regularization techniques. As the authors note, perhaps the most ubiquitous $L_{p}$-norm-like regularizer in use today is weight decay \cite{krogh1991simple}, which is based on an $L_{2}$-norm coupled with a decay/learning rate. Regularization techniques described in this review revolve around quantities computed on model weights independently, typically an $L_{p}$-norm.

Perhaps the most desirable measure of model regularization is the $L_{0}$-norm, which is a count of the number of nonzero parameters in a model. The $L_{0}$-norm is distinguished by the fact that it is scale-invariant (insensitive to a rescaling of the model weights), and thereby does not impose shrinkage on the weights or conflict with techniques such as batch normalization, but the $L_{0}$-norm is not differentiable. The problem of minimizing the $L_{0}$-norm, however, is NP-hard \cite{ge2011note}, as is any $L_{p}$ norm where $p < 1$. A variety of techniques have been suggested to aid in optimizing the $L_{0}$-norm, such as the relaxation achieved by using stochastic gates \cite{louizos2017learning}, or constraint-based optimization of that relaxation \cite{gallego2022controlled}.

Difficulties associated with optimizing the $L_{0}$-norm lead to consideration of regularization based on the $L_{1}$-norm and $L_{2}$-norm, which are differentiable, but come with the cost of imposing an undesirable decrease in the model weights--a phenomenon known as shrinkage--and require the introduction of a regularization parameter which can be difficult to optimize. One recent technique uses a ratio of the square of the $L_{1}$-norm to the $L_{2}$-norm to provide a differentiable scale-invariant regularizer \cite{Yang2020DeepHoyer}. 

This paper proposes a novel regularizer based on the introduction of an additive term to the loss function which incentivizes localization of weights within each layer. This regularizer is motivated by a novel method of estimating entropy and seeks to minimize the entropy of the model given this measure. The resulting regularizer is scale-invariant and therefore should not impose shrinkage on the model weights. This regularizer is differentiable and highly efficient in both memory and compute requirements. Conceptually, this regularizer is accompanied by a shift from thinking about the number of \emph{model parameters} to an estimate of \emph{model entropy}.

\section{Relevant Work}

\paragraph{Total variation regularization} As described in \cite{van2020logistic} and \cite{tibshirani2014adaptive}, total variation regularization utilizes the spatial structure inherent to the outputs of a model via a penalty constructed from a difference between neighboring model output values, but TV regularization does not compute a quantity which is scale-invariant and it does not directly impose sparsity on weights, unlike $L_{p}$-norm-like regularization. Total variation regularization is often described as a so-called ``anisotropic'' regularizer which calculates absolute values of differences between neighboring values of the output of a model, although some sources describe ``isotropic'' calculations based on various discretizations of numeric gradient operators \cite{ortelli2020adaptive}.  \cite{yeh2022total} describe differentiable total variation layers, wherein the layer itself computes an approximation of the layer input which minimizes a total variation penalty. \cite{zhou2004regularization} describe a total variation regularizer for general graph neural networks. The novel sparse regularizer introduced in this paper differs from total variation regularization in that it is scale-invariant due to a weight magnitude normalization term and the novel regularizer is applied throughout the layer weights rather than only to the final model output. 

\paragraph{Graph sparsity} \cite{yu2022principle} define a graph sparsification objective function with the intention of identifying surrogate subgraphs from an the input graph with lower edge count but not fewer graph nodes. The objective function quantifies a divergence between the surrogate and input graphs via a so-called von Neumann graph entropy \cite{passerini2008neumann} computed on trace-normalized graph Laplacians. As noted in the paper, the objective function requires a costly eigenvalue decomposition, for which an approximation of the regularization is offered in the form of a Shannon discrete entropy computed on the normalized degree of nodes. Although the regularizer in the paper is described as differentiable, the differentiability is not intrinsic to the regularizer but added after the fact via introduction of a parameterized stochastic process similar in nature to the stochastic gating as applied to $L_{0}$ regularization by \cite{louizos2017learning}. The graph von Neumann entropy used in the paper is originally described in \cite{minello2019neumann}, which defines both a Laplacian entropy and normalized Laplacian entropy on graphs. Although the graph von Neumann entropy follows the formula for the von Neumann entropy, the paper states there is no known explanation what the quantity means due to the need to estimate probabilities from many observations. The paper treats the graph structure as a discrete entity which, like the $L_{0}$-norm, presents difficulty for optimization via the described objective function, as discussed in \cite{braunstein2006some}.

\paragraph{Fiedler values} \cite{tam2020fiedler} describe a sparse regularizer based on an approximate Fiedler value for graph neural networks, but the technique is not scale invariant and is burdened by an expensive eigenvalue computation. The paper provides a remedy for the compute cost by computing a variational upper bound approximation of the Fiedler value and by not updating the regularization term with each iteration. The paper does point out, however, that the underlying structure of the model (e.g. neural network) provides connectivity information which they consider to be valuable for regularization. The regularizer described in the paper is a ``structurally weighted" $L_1$-like penalty computed via a product of the absolute value of each weight with the square of the difference of approximate Fiedler vectors for neighboring nodes.

\section{Model Entropy}

Although regularization ordinarily involves consideration of the number of model parameters or their magnitudes, the true objective of sparse regularization is to reduce \emph{model entropy}, an estimate of the amount of model information. It is impractical to compute the model entropy from a direct interpretation of the Shannon entropy definition due to the very large number of observations of model parameters which would be required to provide good estimates of a model whose parameters constitute, for instance, a 256x256 weight matrix. Furthermore, this unwieldy estimate would need to be computed regularly and rapidly for the loss function in order to provide a gradient signal for updating the model parameters, and therefore it is necessary to have a model entropy estimate which can be computed accurately and efficiently. The purpose of the paper is to provide this entropy estimate.

\subsection{The blindness of the $L_{p}$-norm to matrix sparsity}

\begin{figure}[h]
	
	\begin{center}
		\begin{minipage}{0.45\textwidth}
			\begin{align*}
				\fbox{\includegraphics[scale=0.5]{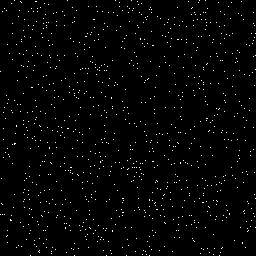}}
			\end{align*}
		\end{minipage}
		\begin{minipage}{0.45\textwidth}
			\begin{align*}
				\fbox{\includegraphics[scale=0.5]{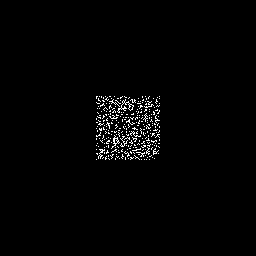}}
			\end{align*}
		\end{minipage}
	\end{center}
	\caption{On the left is a typical example of $W_D$, a dense high-entropy weight matrix and on the right is a typical example of $W_S$, a sparse low-entropy matrix. $W_D$ and $W_S$ are each of dimension 256x256 and each have 1024 entries of value 1 depicted in white (zero entries are black). In $W_D$, the entries are permuted uniformly at random within the entirety of the 256x256 weight matrix. In $W_S$, the value 1 entries are confined to the center 64x64 region, and only that center region is permuted uniformly at random, while values outside of the center region are set to zero. Despite the fact that the human eye can easily see that $W_D$ has higher entropy than $W_S$ and conversely that $W_S$ is more sparse than $W_D$, the $L_p$-norm of both of these matrices are the same for all values of $p$!}	\label{fig:wdws}
\end{figure}

Let us consider two 256x256 matrices where both have 1024 non-zero parameters and each parameter is either precisely 0 or 1: a dense weight matrix $W_D$ and a sparse weight matrix $W_S$. The locations of weight values for $W_D$ may be randomly situated so as to be equally probable anywhere within the 256x256 weight matrix, whereas the weight values for $W_S$ are only selected from the middle 64x64 region, as illustrated in Figure \ref{fig:wdws}.

We would like to compute an estimate of the Shannon entropy of each weight matrix $H(W_D)$ and $H(W_S)$. The Shannon entropy $H(\cdot)$ of a weight matrix $W$ is the sum of three terms:

\begin{equation}
	\label{eqn:fullh}
	H(W) = H_{P}(W) + H_{C}(W) - I_{C;P}(W)
\end{equation}

The first term, the \emph{parameter entropy}, $H_P(W)$ is the contribution of each weight parameter in isolation to the entropy of model weights $H(W)$. In this discussion we will set $H_P(W)$ aside from consideration temporarily by using a binary-valued matrix: each parameter within $W_D$ and $W_S$ may have only values 0 or 1 and therefore have at most one bit of entropy per parameter. The third term, $I_{C;P}(W)$ is the mutual information between the parameter entropy and their sparsity. Because we are fixing each parameter to be one bit, in this discussion we may also drop $I_{C;P}(W)$ from consideration. In a more general setting, we may drop the $I_{C;P}(W)$ term and say that we are optimizing an upper bound on $H(W)$. 

The second term and the primary emphasis of this paper is the \emph{configuration entropy} $H_C(W)$, which is the model weight matrix entropy ignoring this individual per-parameter contribution. $H_C(W)$ increases when weights may be nonzero in more locations within $W$.

%$H_C(W)$ does not increase when there are more nonzero parameters. %

The central contribution of this paper is an estimate for $H_C(W)$, which shall be written as $\ell_{C}(W)$. The entropy of the full model $H(W)$ is upper-bounded by the sum of the entropy of each layer weight matrix individually; we use the approximation that the mutual information of the weight parameter and configuration entropy within a given layer and across layers is zero:

\begin{equation}
	\label{eqn:hdropi}
	H(W) \leq \sum_{i} H_{C}(W_i) + H_{P}(W_i)
\end{equation}

Although it is possible in general that there is nonzero mutual information between the parameter and configurational entropy within and across layers, we are seeking to minimize $H(W)$ and therefore this quantity is useful as a simple upper bound on what would otherwise be a tighter estimate of the true model entropy. It is rare for models to have the necessary structure to represent nonzero $I_{C;P}(W)$, therefore the upper bound may be taken as an approximate equality in practice.

Our purpose in using a binary matrix is to simplify the discussion and assign equal entropy per-parameter irrespective of configuration, when in the more general case the independent entropy contribution from each parameter may vary. If we compute the $L_{p}$-norm for both matrices in Figure \ref{fig:wdws} with the same value of $p$, we will find that $||W_D||_{p} = ||W_S||_{p} = 1024$, which is fixed irrespective of the configuration of the weights, and thus the configuration entropy is unavailable for optimization via  $L_{p}$-norm. Note that $L_{p}$-norm-like regularizers can only optimize $H_{P}(W_i)$.

$L_p$-norm-like regularizers are based on the number of parameters or their magnitudes, but are insensitive to the arrangement of the weight values within a given weight matrix--that is to say, $L_p$-norm-like regularizers are insensitive to permutations of weights within or across weight matrices.

For $L_{p \neq 0}$-norms where $p$ is nonzero such as the $L_1$ and $L_2$-norm, optimizing the $L_{p \neq 0}$-norm will reduce the number of bits transmitted from layer to layer. Batch-normalization counteracts this process, increasing the likelihood that bits are transmitted from one layer to the next. If we consider a deep learning model consisting of weight matrices separated only by a rectified linear unit, we can see that the $L_{p \neq 0}$-norm minimizes the weight values and thereby reduces the likelihood that the input to that layer crosses a threshold in a subsequent layer, thus reducing the number of bits transmitted from one layer to the next. Therefore, when $p$ is nonzero $L_p$-norm-based regularization acts to decrease $H_P(W)$. 

\section{Novel Sparsity Loss}

Our primary object of concern for this new sparse regularizer is the weight matrices themselves, and in particular the extent to which weights are spread uniformly or non-uniformly across the weight matrices. We are not interested to compute the number of nonzero weights as is computed by the $L_0$-norm, nor are we interested to compute a measure of the combined magnitude of the weights, as is the concern of the $L_1$ and $L_2$ norm.

Due to the fact that we need to carefully distinguish between the weight gradient with respect to the loss function and the gradient of the weights as one moves across the weight matrix, this paper will introduce a notation to highlight this distinction, which depicts a square box subscript for the gradient symbol to remind the reader that the subject of this gradient operation is the weight matrix itself (which is shaped like a box):  ${\nabla_\square} W$. 
% {{\overset{\square}{\nabla}} W} -- alternatively over the nabla$

The simplest definition of the weight matrix gradient ${\nabla_\square} W$ would be a pair of matrices: the difference between neighboring weights in the row direction and the differences between neighboring weights in the column direction. This definition is not unique, however, as there are many ways of defining a discrete gradient operator with various trade-offs in symmetry and implementation. In this paper we use normalized Scharr operators $\textbf{G}_{x}$ and $\textbf{G}_{y}$ \cite{scharr2000optimal} rather than using a simple neighbor difference operator so as to maximize rotational symmetry in the resulting weight matrix gradient, although this precaution may not be strictly necessary. In the below, $\star$ denotes two-dimensional convolution with a zero-padded matrix, and computing ${\nabla_\square} W$ produces two matrices, an x-gradient and a y-gradient corresponding to the row and column directions of $W$ respectively.

\begin{minipage}{0.3\textwidth}
	\begin{align*}
		{\nabla_\square} W = (\textbf{G}_{x} \star W, \textbf{G}_{y} \star W)
	\end{align*}
\end{minipage}
\begin{minipage}{0.3\textwidth}
	\begin{align*}
		\textbf{G}_{x} = 	\frac{1}{32}
		\begin{bmatrix}  
			3 & 0 &  -3\\
			10 & 0 & -10\\
			3 & 0 &  -3\\
		\end{bmatrix}
	\end{align*}
\end{minipage}
\begin{minipage}{0.3\textwidth}
	\begin{align*}
		\textbf{G}_{y} = 	\frac{1}{32}
		\begin{bmatrix}  
			3 &  10 & 3\\
			0 &   0 & 0\\
			-3 & -10 &-3\\
		\end{bmatrix}
	\end{align*}
\end{minipage}

To calculate the sparsity loss $\ell_{C}(W)$ we are interested in the magnitude of these matrix gradient parameters, but we are not interested in the overall scale of the parameters of $W$. Therefore we seek to compute the square root of the sum of one half of the element-wise squares of the matrix gradients, but then we wish to divide this quantity by the sum of the absolute values of $W$. Because the square root and division operations are costly to compute and make optimization difficult, we instead compute a logarithm of this overall quantity, yielding the definition below. The \emph{negative} novel sparsity loss for a given weight matrix is then given by:

\begin{equation}
	-\ell_{C}(W) = \frac{1}{2} \log( \frac{1}{2}\sum {\nabla_\square W} \cdot {\nabla_\square W} )  - \log (\sum |W|) = H_{C}(W)
\end{equation} 

The reader should please be careful to note for any independent replication of these results that the right hand side defines the \emph{negative} of the sparsity loss $\ell_{C}(W)$, as a \emph{more positive} value when the right hand side is \emph{less sparse}. The one-half within the logarithm term comes from the fact that $W$ is rank-2 and therefore there are two components to ${\nabla_\square} W$. Although here we discuss weights residing in rank-2 tensors, this novel sparse regularizer should be considered to be defined for any tensor rank via an appropriate higher rank discretized gradient (e.g. Scharr) operator and normalization.

As the reader can verify, $\ell_{C}$ is scale invariant: $\ell_{C}(W) = \ell_{C}(\alpha W)$ for scalar $\alpha$. This can be seen either by considering that the log of a product is the sum of the logs and that the two terms within $\ell_{C}$ cancel, or by considering that the original construction was a ratio of terms proportionate to $W$. Scale invariance is of particular importance in the context of deep neural networks as the shrinkage (i.e. lack of scale-invariance) in $L_{p \neq 0}$-norm regularization directly conflicts with batch normalization \cite{azarian2020learned}.

It is intuitively clear from visual inspection of Figure \ref{fig:wdws} that the entropy of $W_D$ is large whereas the entropy of $W_S$ is much lower, or equivalently that $W_D$ is dense whereas $W_S$ is sparse. Recall that the Shannon entropy can be formulated as an expectation value over the probabilities of each state: 

\begin{equation}
	H = \sum - p_{i} \log p_{i} = \E [- \log p_{i}]
\end{equation}

Although frequently referred to as ``the'' entropy, Shannon entropy is one of a variety of estimates of the amount of information in a variable, but there are many other methods of carrying out this estimate including, for example, Tsallis entropy \cite{tsallis1988possible}, the Rényi entropy \cite{renyi1961measures}, and the Hartley entropy \cite{hartley1928transmission}. Like the $L_{p}$-norm, these entropy measures are calculated on variables in isolation from one another. This paper is defining what appears to be a new measure of entropy wherein the variables of interest carry a spatial arrangement such as a matrix or tensor, the measure examines the spatial arrangement of the variables, and where greater localization corresponds to lower entropy.

This visual intuition that we may confidently state that the entropy of $W_D$ is large whereas the entropy of $W_S$ is small is a direct manifestation of the asymptotic equipartition property, as the individual observations of $W_D$ and $W_S$ can be considered to be almost surely part of the typical set, with tighter bounds as the number of variables and as the available empty space in $W$ increases. If we consider small subregions of a given matrix, subregions which are zero do not contribute to $\ell_{C}(W)$. Subregions having a larger boundary (greater spacing between entries of value one) produce a larger contribution to $\ell_{C}(W)$ than subregions having a smaller boundary. If we consider gradually squishing together the  entries of value one in a matrix of high or maximal configuration entropy such as $W_D$, the probability that subregions will have a smaller boundary goes up while the number of subregions which are zero also increases, and thus  $\ell_{C}(W)$ decreases. The sparsity loss $\ell_{C}(W)$ can be interpreted as an estimate of the configuration entropy $H_C(W)$, which converges more tightly as the dimension of $W$ increases and as the number of observations increases. 

\subsection{Connection to the Helmholtz Equation and Rayleigh Quotient}

Although $\ell_{C}$ was originally derived based on intuition and qualitative judgments about the desired characteristics of a regularizer, further inspection reveals connections between this quantity, the Helmholtz equation defined on the weight matrix, and the corresponding Rayleigh quotient. In particular, $\ell_{C}$ is a logarithm of a discretization of the Rayleigh quotient when we are solving the Helmholtz equation on the weight matrix (treating the rows and columns of the weight matrix as the spatial dimensions of the differential equation). The reader may recognize the Helmholtz equation as the time-independent Schrodinger equation from quantum mechanics and the Rayleigh quotient as the energy computed for a specific proposal wavefunction in the variational method energy functional. Remarkably, finding an accurate, low entropy model is equivalent to finding the ground state of a molecule and its electronic wavefunction!

\begin{equation}
	\begin{cases}
		\nabla^2_\square W = \lambda W & \text{on } \Omega \\
		W = 0 &  \text{on } \partial \Omega \\
	\end{cases}
\end{equation}

\begin{equation}
	\label{eqn:minrq}
	\begin{aligned}
		\underset{W} \min \frac{\int_\Omega ||\nabla_\square W||^2}{ \int_\Omega ||W||^2} & = \underset{W} \min \log \int_\Omega ||\nabla_\square W||^2 - \log \int_\Omega ||W||^2 \\
		& = \underset{W} \min \frac{1}{2} \log \int_\Omega ||\nabla_\square W \cdot \nabla_\square W || - \log \int_\Omega ||W|| \\
		& \approx \underset{W} \min \frac{1}{2} \log( \frac{1}{2}\sum {\nabla_\square W} \cdot {{\nabla_\square} W} )  - \log (\sum |W|) 
	\end{aligned}
\end{equation}

Whereas for the variational ansatz the Rayleigh quotient is minimized to determine a ground state, this may be interpreted as a joint optimization of both the wavefunction and the ground state energy itself. Note that an alternative definition of $\ell_{C}$ may make use of the left hand side of Equation \ref{eqn:minrq}.

\section{Parameter Entropy}

In a general setting, the mutual information of the parameter and configuration entropy may not be zero, and so therefore the minimization of the novel sparsity loss may be thought of as minimizing an upper bound estimate of the model configuration entropy. This paper does not optimize the parameter entropy, but such a method could be constructed from a stochastic gating technique truncating the input to a certain number of bits (including to zero bits). $L_{p}$-norm-like constructs can be viewed as seeking to reduce the parameter entropy by increasing the probability that input bits end up truncated by a subsequent nonlinearity. In the case of the $L_{0}$-norm, the entropy per parameter may be regarded as fixed, often as a 16, 32, or 64-bit floating point number.

\section{Experimental results}

\begin{figure}
	\includegraphics[width=\textwidth]{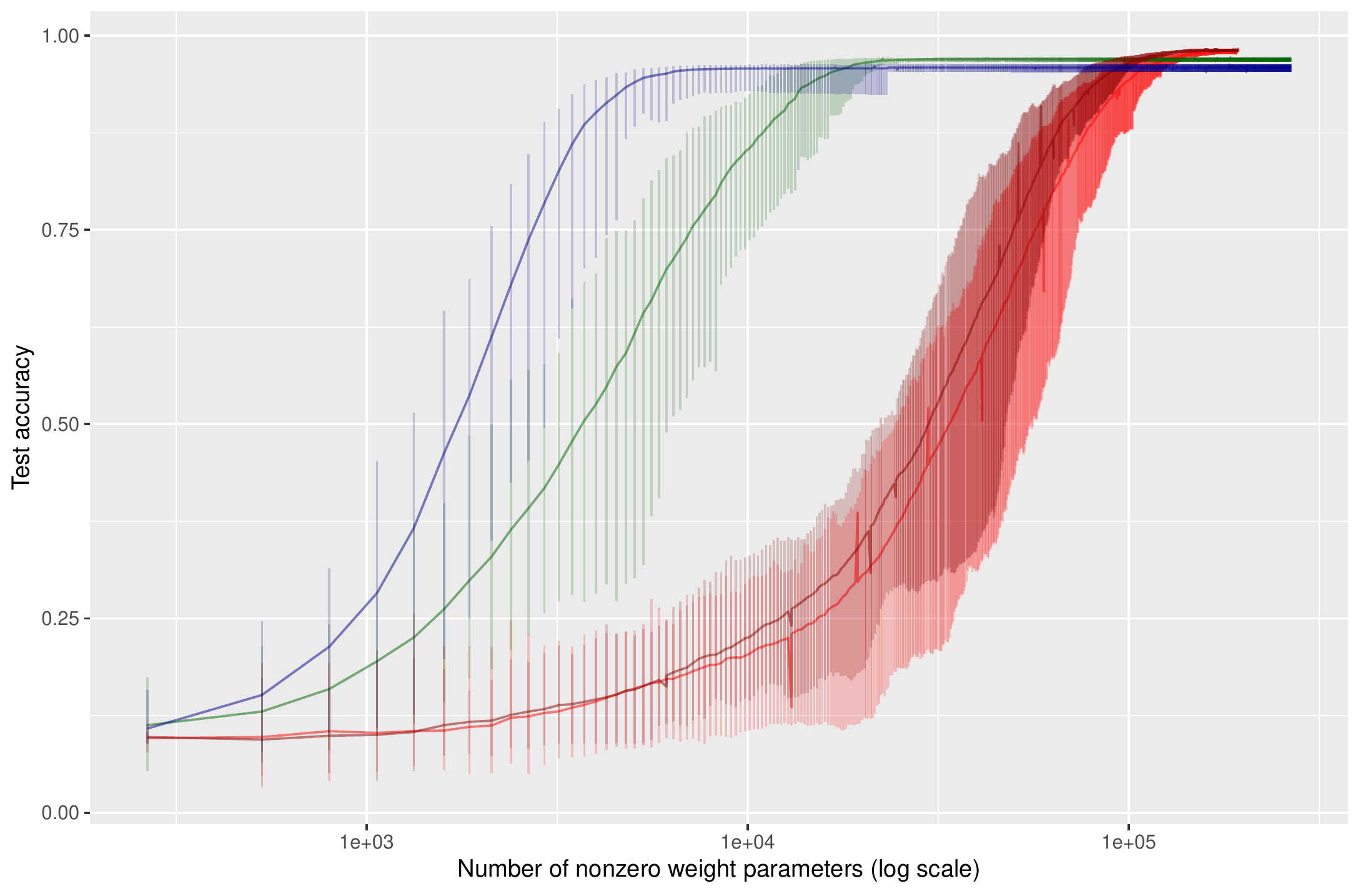}
	\caption{This figure depicts model performance versus parameter count when the weights are thresholded at a variety of parameter counts based on their sensitivity (product of weight magnitude with weight gradient). The model used was LeNet300, which was trained on MNIST using a cross entropy loss without the novel sparsity loss and without weight decay (red), without the novel sparsity loss and with weight decay (dark red), with the novel sparsity loss and no weight decay (green), and with the novel sparsity loss and with weight decay (blue). Each model was retrained 64 times with different starting seeds, and the thin vertical lines represent the minimum and maximum accuracy across all 64 training runs with the centerline representing a median accuracy for a given number of parameters. In all cases where weight decay was used, the weight decay parameter was 1e-4. The sharp excursions in the red and dark red lines do not appear to be any form of implementation error and instead likely reflect a greater discrepancy between the sensitivity percentile and the actual impact on accuracy of removing parameters at around that percentile.  } \label{fig:tap}
\end{figure}

LeNet300 is a simple multilayer perceptron with layer architecture 784-300-100, and in these experiments this model was used with a regularized linear unit as the nonlinearity.  The LeNet300 model was trained on MNIST \cite{lecun1998gradient} using a cross entropy loss without the novel sparsity loss and without weight decay (red), without the novel sparsity loss and with weight decay (dark red), with the novel sparsity loss and no weight decay (green), and with the novel sparsity loss and with weight decay (blue), with results depicted in Figure \ref{fig:tap}. In all cases the number of training epochs was the same, and the Adam optimizer was used with a learning rate of 1e-4. Where weight decay was used, it was assigned a parameter value of 1e-4. The machine used to perform this experimentation was a quad-core Intel CPU with an NVIDIA GeForce GTX 1060 6GB GPU running CUDA 12.0.76 with libtorch 1.13.0. The solid line in Figure \ref{fig:tap} represents the median of 64 distinctly seeded training runs, and the vertical lines represent the full range between the highest and lowest test accuracy across all runs. 

To evaluate the model after training, the sensitivity of each weight was calculated, where sensitivity is defined as the absolute value of the product of a weight with the (traditional) weight gradient. The sensitivity was computed across all weights and layers and thresholded at 0.1 percentile increments. For each 0.1 percentile sensitivity increment, the test accuracy and number of weights above the sensitivity threshold was computed. 

At all accuracy levels the model optimized with the novel sparse loss had an approximately one order of magnitude advantage in number of parameters required to achieve a given accuracy (Figure \ref{fig:tap}). For example, with 19,166 parameters one sparsely trained model without weight decay achieved a 96.01 percent accuracy while a densely trained model without weight decay achieved a 96.03 percent accuracy with 127,776 parameters. Where both weight decay and the novel sparsity loss were used, as compared to when neither were used, the parameter count advantage at high accuracy approached a factor of fifty .

Note that a log scale is used for the horizontal axis in Figure \ref{fig:tap}, which represents the number of parameters after pruning based on a sensitivity threshold. Also note that the accuracy for a given parameter count is determined only by sensitivity-based thresholding of the parameters resulting from the single model training. Each of the 64 training runs produces one accuracy-parameter curve, and each training run was simply stopped at one hundred epochs.  The model resulting from having both the novel sparse loss and the weight decay was considerably more sparse than either independently, which may be interpreted as an empirical manifestation of the separability of parameter and configuration entropy within the model and the validity of optimizing the upper bound on $H(W)$ as a technique for improving model sparsity. That is to say, these results experimentally suggest that we may reasonably drop the $I_{C;P}(W)$ term from the full entropy in Equation \ref{eqn:fullh} and still obtain good results from optimizing the upper bound in Equation \ref{eqn:hdropi}.

The gradient was observed to rapidly becomes sparse during training, suggesting that this technique could be used to generate a faster (sparse) training method, possibly with an initially slow and dense initialization.

In contrast to traditional $L_{p}$-based regularization schemes, no regularization parameter was applied to the sparse loss term, nor was hyperparameter tuning performed. This may be because the units of the model error (cross entropy in nats) are well matched to the units of the novel sparsity loss (model entropy in nats). In practice it may make sense to introduce a regularization parameter or hyperparameter, possibly on a per-layer basis.

Inspection of the densely trained median test accuracy in Figure \ref{fig:tap} reveals  sharp and significant excursions in test accuracy as the sensitivity threshold is steadily decreased throughout a wide range of thresholds. Very careful inspection of the sparsely trained median test accuracy at higher accuracy values also reveals very small sharp excursions in test accuracy as the sensitivity threshold is decreased, but these small excursions are almost entirely concentrated in the least important weights, as predicted by sensitivity percentile. None of these sharp excursions appear to be an implementation error: the same code was used to generate all four plots. The only difference between the sparse and dense plots was that the novel sparsity loss was added to the loss function. Instead, these excursions may be interpreted as the novel sparse regularization term both making it easier to predict which weights can be removed via sensitivity and lessening the impact of removal of unimportant weights. Thus, not only does the novel sparse regularizer make the model more sparse in a continuous sense (e.g. pulling unimportant weights towards zero) but it is also easier to accurately identify the order of weight importance for model accuracy via sensitivity, and therefore the novel sparse regularizer also renders the model more absolutely sparse in a discrete sense as the parameters are removed.

\subsection{Implementation}

Wall-clock training time ranged from 76 seconds without weight decay or the novel sparsity loss to 285 seconds with weight decay and the novel sparsity loss across all training runs and conditions. This implementation, which was not well-optimized, allocated additional weight matrices for the convolution operations and therefore used a few times the memory of the dense model during training, but note that a high-quality implementation of this technique would make use of a fused convolution-accumulation operation to introduce extraordinarily small memory overhead (a small number of bytes per layer) to compute the novel sparsity loss. Also note that this loss function is quite trivial to parallelize. There are some variations on the potential choice of gradient operator discretization, some of which involve a smaller kernel such as a simple neighbor subtraction, but the Scharr operator was chosen so as to minimize any induced angular bias in the resulting computation.

\newpage
\begin{figure}[h]
	
	\begin{center}
		\begin{minipage}{0.45\textwidth}
			\begin{align*}
				\fbox{\includegraphics[scale=0.5]{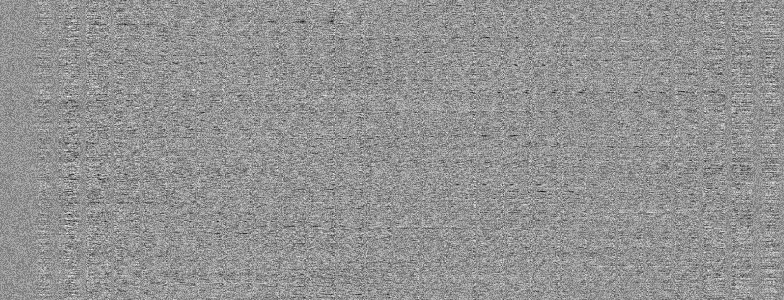}}
			\end{align*}
		\end{minipage}
		\begin{minipage}{0.45\textwidth}
			\begin{align*}
				\fbox{\includegraphics[scale=0.5]{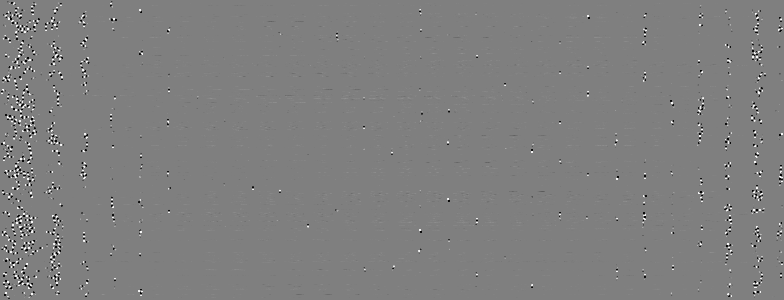}}
			\end{align*}
		\end{minipage}
	\end{center}
	\begin{center}
		\begin{minipage}{0.45\textwidth}
			\begin{align*}
				\fbox{\includegraphics[scale=0.5]{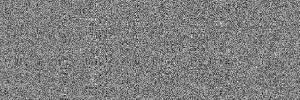}}
			\end{align*}
		\end{minipage}
		\begin{minipage}{0.45\textwidth}
			\begin{align*}
				\fbox{\includegraphics[scale=0.5]{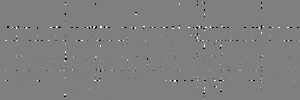}}
			\end{align*}
		\end{minipage}
	\end{center}
	\begin{center}
		\begin{minipage}{0.45\textwidth}
			\begin{align*}
				\fbox{\includegraphics[scale=1]{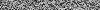}}
			\end{align*}
		\end{minipage}
		\begin{minipage}{0.45\textwidth}
			\begin{align*}
				\fbox{\includegraphics[scale=1]{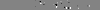}}
			\end{align*}
		\end{minipage}
	\end{center}
	\caption{On the top (large 300x784 matrix) or left (smaller 300x100 and 10x100 matrices) we have trained weights for the LeNet300 model on MNIST using a traditional dense method (cross entropy loss only and without weight decay). Grey values are zero, white are positive, and black are negative. On the bottom or right we can see the weights produced by the sum of the traditional cross entropy loss and the novel sparsity loss using the same model on the same data without weight decay. Note that the weights on the bottom or right are depicted as they are immediately at the conclusion of training: no pruing or thresholding was applied to the weights in this Figure. }	\label{fig:wdvss}
\end{figure}

\newpage

\newpage
\section{Conclusion}
Although traditionally, information is viewed as residing in an ideal platonic void, this paper explores the implications of information having a physical location in space. The most significant such implication is that a novel definition of entropy may be constructed based on the localization of information, with higher entropy associated with a more uniform distribution of that information. This novel entropy is then used as an estimate of model entropy for neural networks and minimized as a sparse regularizer when added to the traditional cross-entropy loss during training. This novel regularizer is scale-invariant, differentiable, and can be efficiently computed with trivial additional memory and compute. Adding this regularizer to the common cross-entropy loss achieves approximately a one order of magnitude improvement in the number of parameters required to achieve a given level of accuracy across a wide range of accuracies for a LeNet300 model trained on MNIST.

\bibliographystyle{plainnat}
\bibliography{neurips2023_sparse_loss_references}

%%%%%%%%%%%%%%%%%%%%%%%%%%%%%%%%%%%%%%%%%%%%%%%%%%%%%%%%%%%%

\end{document}